\pdfoutput=1

\documentclass[11pt]{article}

\usepackage[final]{acl}

\usepackage{times}
\usepackage{latexsym}

\usepackage[T1]{fontenc}

\usepackage[utf8]{inputenc}

\usepackage{microtype}

\usepackage{inconsolata}

\usepackage{graphicx}

\usepackage{amssymb, amsmath}
\usepackage{siunitx}
\usepackage{bm}
\usepackage{pifont}
\definecolor{Gray}{gray}{0.9}
\definecolor{gr}{RGB}{0, 146, 0}

\newcommand{\org}[1]{\textcolor{orange}{#1}}
\newcommand{\red}[1]{\textcolor{lightred}{#1}}

\definecolor{orange}{HTML}{ff9900} 

\definecolor{green}{HTML}{34a853}
\definecolor{lightgreen}{HTML}{b6d7a8}
\definecolor{seagreen}{HTML}{3CB371}

\definecolor{lightgray1}{HTML}{d9d9d9}
\definecolor{lightyellow2}{HTML}{ffe599}

\definecolor{blue}{HTML}{4285f4}
\definecolor{lightblue}{HTML}{9fc5e8}
\definecolor{lightcornflowerblue3}{HTML}{c9daf8}

\definecolor{purple}{HTML}{9900ff} 
\definecolor{lightpurple1}{HTML}{8e7cc3}
\definecolor{lightpurple}{HTML}{b4a7d6}
\definecolor{lightred}{HTML}{e06666}

\usepackage{soul}
\colorlet{exqcolor}{lightgreen!70}
\colorlet{exccolor}{lightgray1!55}
\colorlet{exdcolor}{orange!30}
\colorlet{exacolor}{lightblue!90}
\colorlet{execolor}{lightred!50}
\colorlet{exncolor}{lightpurple!50}

\newcommand{\exd}[1]{\sethlcolor{exdcolor}\hl{#1}}

\newcommand{\exe}[1]{\sethlcolor{execolor}\hl{#1}}

\usepackage{multirow}
\usepackage{booktabs}
\usepackage{graphicx}
\usepackage{tabularx}
\usepackage{adjustbox}
\usepackage{array}
\usepackage{pifont}
\usepackage{xspace}
\usepackage{tcolorbox}
\usepackage{colortbl}
\usepackage{hyperref}
\usepackage{arydshln}
\usepackage{wrapfig}
\usepackage{soul}
\usepackage{makecell}
\usepackage{color, xcolor}
\usepackage{pgfplots}
\definecolor{tiffanyblue}{RGB}{129,216,208}
\definecolor{bangdiblue}{RGB}{0,149,182}
\definecolor{kleinblue}{RGB}{0,47,167}
\usepackage{amssymb}
\usepackage{circledtext}
\usetikzlibrary{shapes}
\usetikzlibrary{arrows,decorations.pathmorphing,backgrounds,positioning,fit,petri}
\usetikzlibrary{arrows.meta,fit,shapes.arrows}
\usetikzlibrary{positioning,shadows,patterns}
\usepackage{caption} 
\usepackage{tikz}
\usepackage{subfig}
\usepackage{tkz-kiviat,pgfplots}
\usepackage{xspace}
\newcommand{\posi}{{\tt[POS]}\xspace}
\newcommand{\nega}{{\tt[NEG]}\xspace}
\def\yPos{y^+}
\def\yNeg{y^-}
\newcommand{\ours}{\textsc{Rhio}\xspace}
\newcommand{\benchmark}{GroundBench\xspace}
\newcommand{\eg}{\emph{e.g.}}

\makeatletter
\def\adl@drawiv#1#2#3{%
        \hskip.5\tabcolsep
        \xleaders#3{#2.5\@tempdimb #1{1}#2.5\@tempdimb}%
                #2\z@ plus1fil minus1fil\relax
        \hskip.5\tabcolsep}
        
\newcommand{\cdashlinelr}[1]{%
  \noalign{\vskip\aboverulesep
           \global\let\@dashdrawstore\adl@draw
           \global\let\adl@draw\adl@drawiv}
  \cdashline{#1}
  \noalign{\global\let\adl@draw\@dashdrawstore
           \vskip\belowrulesep}}
\makeatother

\title{Improving Contextual Faithfulness of Large Language Models via \\  Retrieval Heads-Induced Optimization}

\author{
 \textbf{Lei Huang\textsuperscript{1}},
 \textbf{Xiaocheng Feng\textsuperscript{1,2}\thanks{Corresponding Author}},
 \textbf{Weitao Ma\textsuperscript{1}},
 \textbf{Yuchun Fan\textsuperscript{3}},
 \textbf{Xiachong Feng\textsuperscript{4}},
\textbf{Yangfan Ye\textsuperscript{1}},
\\
 \textbf{Weihong Zhong\textsuperscript{1}},
 \textbf{Yuxuan Gu\textsuperscript{1}},
 \textbf{Baoxin Wang\textsuperscript{5}},
 \textbf{Dayong Wu\textsuperscript{5}},
 \textbf{Guoping Hu\textsuperscript{5}},
 \textbf{Bing Qin\textsuperscript{1,2}}
\\
 \textsuperscript{1} Harbin Institute of Technology, China
 \textsuperscript{2} Peng Cheng Laboratory, China 
 \\
 \textsuperscript{3} Northeastern University, China
 \textsuperscript{4} The University of Hong Kong, China
 \textsuperscript{5} iFLYTEK Research, China
 \\
 \texttt{\{lhuang, xcfeng, wtma, yfye whzhong, yxgu, qinb\}@ir.hit.edu.cn}
 \\
 \texttt{yuchunfan\_neu@outlook.com} \quad
  \texttt{fengxc@hku.hk} \quad
 \texttt{\{bxwang2, dywu2, gphu\}@iflytek.com}
}

\begin{document}
\maketitle

\begin{abstract}
Ensuring contextual faithfulness in retrieval-augmented large language models (LLMs) is crucial for building trustworthy information-seeking systems, particularly in long-form question-answering (LFQA) scenarios.
In this work, we identify a salient correlation between LFQA faithfulness and \textit{retrieval heads}, a set of attention heads responsible for retrieving contextual information.
Leveraging this insight, we propose \ours\footnote{Acronym for \textbf{R}etrieval \textbf{H}eads-\textbf{I}nduced \textbf{O}ptimization}, a framework designed to teach LLMs to explicitly discriminate between faithful and unfaithful generations. 
\ours first augments unfaithful samples that simulate realistic model-intrinsic errors by selectively masking retrieval heads. Then, these samples are incorporated into joint training, enabling the model to distinguish unfaithful outputs from faithful ones conditioned on \textit{control tokens}. Furthermore, these control tokens are leveraged to self-induce contrastive outputs, amplifying their difference through contrastive decoding.
Additionally, to facilitate the evaluation of contextual faithfulness, we also introduce \benchmark, a comprehensive benchmark compiled from five existing LFQA datasets. 
Extensive experimental results on \benchmark demonstrate that \ours significantly improves faithfulness, even outperforming GPT-4o\footnote{\url{https://github.com/LuckyyySTA/faithful-LFQA}}.
\end{abstract}
\section{Introduction}
Large language models (LLMs) have achieved remarkable advancements across various domains due to their powerful generative capabilities~\citep{touvron2023llama2, openai2023gpt4}. 
By integrating retrieval techniques, retrieval-augmented LLMs further enhance their performance in information-seeking tasks, producing highly coherent and informative responses~\citep{gao2023retrieval, asai2024reliable}.
Particularly, long-form question answering (LFQA)~\citep{fan2019eli5, stelmakh2022asqa}, aiming at synthesizing information from disparate retrieval sources to provide in-depth, paragraph-length responses, has emerged as a research hotspot due to its wide-ranging application.
However, recent studies~\citep{chen2023understanding, alessandro2024groundedness} reveal that these LLMs suffer from \textit{faithfulness hallucination} in LFQA, where the generated response is not grounded in provided sources~\citep{huang2023a}, as shown in Figure~\ref{tab:example_lfqa}. 
Such unsupported information can significantly erode user trust and undermine the reliability of LLMs.

\begin{table}[t!]
\centering
\scriptsize
\begin{tabular}{p{0.93\columnwidth}}
\toprule
\rowcolor{Gray}
\multicolumn{1}{c}{\textbf{Model Input}}\\
\midrule
\textbf{Question:}  Difference between Russian Blue and British Blue cat? \\
\textbf{Document [1]:} The Russian Blue is a naturally occurring breed that may have originated in the port of Arkhangelsk in Russia. They come in colors varying from a light shimmering silver to a darker, slate grey. (\dots) They are also considered to be hypoallergenic. \\
\textbf{Document [2]:} The coat is known as a ``double coat'', with the undercoat being soft, downy, and equal in length to the guard hairs, which are an even blue with silver tips. (\dots) The coat is described as thick, plush, and soft to the touch. The feeling is softer than the softest silk. \\
\textbf{Document [3]:} The Russian Blue has bright green eyes, pinkish lavender or mauve paws, two layers of short thick fur, and a blue-grey coat. The color is a bluish-gray which is the dilute expression of the black gene. (\dots) Its eyes are almost always a dark and vivid green. Any white patches of fur or yellow eyes in adulthood are seen as flaws in show cats. \\
\midrule
\rowcolor{Gray}
\multicolumn{1}{c}{\textbf{Model Output}}\\
\midrule
\textbf{Unfaithful Response:} The Russian Blue and the British Blue cat differ in several distinct characteristics. One key difference is the length of their coats. While the Russian Blue has a ``double coat'' with a soft, downy undercoat and a plush, soft outer coat, \exe{the British Blue typically has a single coat.} In terms of eye color, Russian Blues typically have dark green eyes, while \exd{British Blues may have yellow eyes or white patches of fur in adulthood, which are considered flaws in show cats.} \\
\bottomrule
\end{tabular}
\caption{An example of the unfaithful response generated by GPT-4o-mini. We highlight the different error types in the response. The content marked in \red{red} is a \textit{fabricated hallucination}, and the content marked in \org{orange} indicates \textit{inaccurate information synthesis}.} 
\label{tab:example_lfqa}
\vspace{-0.45cm}
\end{table} 
Recent efforts to enhance faithfulness in LFQA have primarily focused on improving context quality via explicit denoising \citep{xu2023recomp}, self-critiquing its generation with reflection tokens \citep{asai2024selfrag} and context-aware decoding to amplify contextual information \citep{shi2024trusting}. While effective, these approaches are more of a compensatory way rather than allowing the model to learn to avoid generating unfaithful responses.

In this work, we investigate the idea of explicitly teaching LLMs to discriminate between faithful and unfaithful outputs to improve contextual faithfulness. 
One of the key challenges lies in generating realistic unfaithful samples. 
Inspired by recent research on \textit{retrieval heads} \citep{wu2024retrieval}, a special type of attention head for retrieving information from context, we delve into the role of these retrieval heads in the contextual faithfulness of LFQA. Our pilot study reveals that the activation of retrieval heads potentially explains faithfulness in LFQA: masking out retrieval heads leads to error patterns similar to real unfaithfulness from the model (\S\ref{ssec:retrieval_head}).
This enables us to augment more diverse and realistic unfaithful samples by simply masking out retrieval heads in LLMs (\S\ref{ssec:negative_augmentation}).
Given faithful and augmented unfaithful samples, \ours introduces two special \textit{control tokens}, \posi and \nega, to signal the generation towards either faithful or unfaithful responses. This process supervises the model to explicitly distinguish unfaithful responses from faithful ones (\S\ref{ssec:learning_to_discriminate}).
Furthermore, these control codes are further utilized to induce contrastive generations and amplify the difference between them via contrastive decoding to further enhance faithfulness (\S\ref{ssec:unfaithful_induced_decoding}).

Additionally, to reliably evaluate the faithfulness of LLMs in LFQA, we also introduce \benchmark, a comprehensive benchmark compiled from five existing LFQA datasets. \benchmark is designed to ensure that retrieved documents contain sufficient information to answer questions, thus providing a controlled evaluation setting.
Extensive experiments on \benchmark demonstrate that \ours achieves significant improvements in faithfulness, with average gains of 12.84\% and 12.59\% in 7B and 13B models, respectively, outperforming even the state-of-the-art GPT-4o.
Human evaluation and further analysis reveal additional insights into the efficacy of \ours.
\section{Preliminaries}
\label{sec:preliminary_study}
In this section, we introduce the task formulation of LFQA and our investigation of retrieval heads, which motivates the proposed of \ours.

\subsection{Task Formulation}
The task of LFQA can be described as follows. Given a collection of questions $\mathcal{Q}$ and a corpus of documents $\mathcal{D}$, for a question $q \in \mathcal{Q}$, a retriever $R$ first retrieves $k$ documents, denoted as $R(q) = \{d_1, d_2, \ldots, d_k\}$, where $d_i \in \mathcal{D}$. Subsequently, the question $q$ and the retrieved documents $R(q)$ are combined as the input to the LLM $M$ to generate a paragraph-length response $\mathcal{S} = <s_1, s_2, \ldots>$ consisting of multiple sentences. In this work, we primarily focus on the faithfulness issue in LFQA. Notably, a model-generated response $S$ is considered unfaithful if it contains at least one sentence $s_i \in S$ that either contradicts or cannot be verified using the retrieved document $R(q)$.

\subsection{Retrieval Head}
\label{ssec:retrieval_head}
Recent interpretability research \citep{olsson2022in, wu2024retrieval} has identified a special type of attention head, termed \textit{retrieval heads}, which is largely responsible for retrieving relevant information from contextual sources. When these heads are activated, the model performs a copy-paste operation from the provided context. 
This motivates us to link these retrieval heads with contextual faithfulness in LFQA, which largely relies on the model's ability to synthesize information from contextual cues.
To delve into the implications of retrieval heads in contextual faithfulness, we start with a preliminary study on a long-form QA dataset. 
\definecolor{ublue}{RGB}{52,152,219}
\definecolor{ured}{RGB}{240,100,100}
\definecolor{uorange}{RGB}{247,175,89}
\definecolor{upurple}{RGB}{148,137,250}
\definecolor{pink1}{HTML}{FFF0F6}
\definecolor{pink2}{HTML}{FCC2D7}
\definecolor{pink3}{HTML}{FAA2CA}
\begin{figure}[!t]
\begin{tikzpicture}
\centering
    \scriptsize{
    \begin{axis}[
      at={(0,0)},
      ymajorgrids,
      xmajorgrids,
      grid style=dashed,
      legend style={at={(0.41,1)}, anchor=south west},
      legend cell align={left},
      ybar,
      enlarge x limits=0.15,
      xtick align=inside,
      height=.3\textwidth,
      width=.5\textwidth,
      bar width=1.2em,
      xshift=-1.5em,
    nodes near coords,
    nodes near coords align={vertical},
    nodes near coords style={font=\tiny, scale=0.8,/pgf/number format/fixed, /pgf/number format/precision=1},
     every node near coord/.append style={
        /pgf/number format/.cd,
            fixed,
            fixed zerofill,
            precision=1
    },
      xlabel={\footnotesize Number of masked retrieval heads},
      xmax=4,
      xmin=0,
      symbolic x coords={0,1,2,3,4},
      legend style={cells={align=left}},
      xtick=data,
      nodes near coords align={vertical},
      ymin=30,
      ymax=93,
      ytick={30,40,50,60,70,80,90},
        yticklabels={30.0, 40.0, 50.0, 60.0, 70.0, 80.0, 90.0},
      xticklabels={0, 25, 50, 75, 100},
      xtick style={draw=none},
      ytick style={draw=none},
      ylabel style={yshift=-3em},xlabel style={yshift=0.3em,align=center},
      yticklabel style={/pgf/number format/fixed,/pgf/number format/fixed},
      legend style={draw=none,
        line width=1pt,
        at={(0.5,1.0)},
        anchor=south},
        xtick=data,
        axis on top=false,
      ]
         \addplot[fill=cyan!10,draw=gray, area legend] coordinates {(0,80.1) (1,59.2) (2,48.2) (3,43.1) (4,35.6)};
        \addplot[fill=cyan!40, draw=gray, area legend] coordinates {(0,82.1) (1,69.2) (2,53.6) (3,46.5) (4,41.1)};
          \addplot[ fill=cyan!70,draw=gray, area legend] coordinates {(0,86.4) (1,75.7) (2,70.8) (3,67.6) (4,51.5)};
    \end{axis}
    } 
    \node [rectangle,draw=gray,fill=cyan!10,inner sep=2pt,minimum height=0.8em,minimum width=2.5em,font=\small,anchor=north,align=center,] (label1) at (16em,12em){};
    \node [rectangle,draw=gray,fill=cyan!40,inner sep=2pt,minimum height=0.8em,minimum width=2.5em,font=\small,anchor=north,align=center,] (label2) at (16em,11em){};
    \node [rectangle,draw=gray,,fill=cyan!70,inner sep=2pt,minimum height=0.8em,minimum width=2.5em,font=\small,anchor=north,align=center,] (label3) at (16em,10em){};
    \node [align=center] (label1_1) at ([xshift=4.8em,yshift=-0.35em]label1.north){\tiny Llama-2-7b-chat};
    \node [align=center] (label1_3) at ([xshift=5em,yshift=-0.35em]label2.north){\tiny Llama-2-13b-chat};
    \node [align=center] (label1_2) at ([xshift=5em,yshift=-0.35em]label3.north){\tiny Llama-2-70b-chat};
    \node [rotate=90]at (-4.2em,6.3em) {\footnotesize{FaithScore (\%)}};
\end{tikzpicture}
  \caption{Impact of masking different numbers of masked retrieval heads on model faithfulness.}
  \label{fig:mask_retrieval_head}
  \vspace{-6mm}
\end{figure}
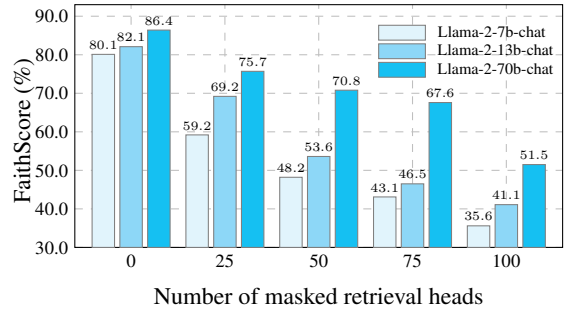
\subsubsection{Experimental Setup for Pilot Study}
We experiment with the CLAPNQ dataset \citep{rosenthal2024clapnq}, a component of our \benchmark, to explore how retrieval heads affect contextual faithfulness in LFQA (for details about CLAPNQ and evaluation of faithfulness, see Section \S\ref{sec:ground_bench}). Specifically, we utilize models from Llama-2 family \citep{touvron2023llama2}, with sizes ranging from 7B to 70B. We first apply the retrieval head detection algorithm proposed by \citet{wu2024retrieval} to detect retrieval heads within each model. Subsequently, We gradually mask out the top retrieval heads, from 0 to 100, to observe the corresponding changes in the faithfulness of the model’s response.
\subsubsection{Observation}
Next, we present our main findings as follows.
\paragraph{Findings 1: Unfaithfulness in LFQA is exacerbated when masking out more retrieval heads.}
As shown in Figure~\ref{fig:mask_retrieval_head}, our results indicate a direct correlation between the number of masked retrieval heads and the severity of unfaithfulness. As the number of masked retrieval heads increases, faithfulness in long-form responses leads to a substantial decrease. In contrast, randomly masking non-retrieval heads shows no significant impact on faithfulness (results available in Appendix~\ref{appendix:random_mask}). 
This observation suggests that retrieval heads are crucial for ensuring contextual faithfulness in LFQA.
\definecolor{border}{HTML}{f7fcfd}
\newcommand{\BorderWidth}{0.5}
\newcommand{\radius}{3}
\newcommand{\InsideTextRadius}{2.3}
\newcommand{\MiddleTextRadius}{1.5}

\begin{figure}[!t]
\centering
\begin{minipage}{.2\textwidth}
  \centering
  \resizebox{\linewidth}{!}{
  \begin{tikzpicture}
  \newcounter{a}
  \newcounter{b}
  \foreach \x/\name/\color/\percentage in {40/Fabricated/61a5c2/40\%, 48/Incomplete/89c2d9/48\%, 12/Inconsistency/a9d6e5/12\%}
  {
      \definecolor{color}{HTML}{\color}
      \setcounter{a}{\value{b}}
      \addtocounter{b}{\x}
      \pgfmathsetmacro{\middleangle}{(\thea+\theb)/2}
      \pgfmathsetmacro{\labelradius}{\InsideTextRadius*0.8}
      \pgfmathsetmacro{\percentageradius}{\MiddleTextRadius}
      \coordinate (target_1) at ({\radius*cos(\thea/100*360)},{\radius*sin(\thea/100*360)});
      \coordinate (label_pos) at ({\labelradius*cos(\middleangle/100*360)}, {\labelradius*sin(\middleangle/100*360)});
      \coordinate (percentage_pos) at ({\labelradius*cos(\middleangle/100*360)}, {\labelradius*sin(\middleangle/100*360)-0.5});
      \filldraw[line width=\BorderWidth mm, draw=border, fill opacity=1, fill=color] (0,0) -- (target_1) arc ({\thea/100*360}: {\theb/100*360} : \radius cm) -- cycle;

      \node at (label_pos) {\textbf{\name}};
      \node at (percentage_pos) {\percentage};
  }
  \end{tikzpicture}}
  \vspace{-20pt}
  \caption*{(a) model-intrinsic}
  \label{fig:subfig_a}
\end{minipage}
\hspace{6mm}
\begin{minipage}{.2\textwidth}
  \centering
  \resizebox{\linewidth}{!}{
  \begin{tikzpicture}
  \newcounter{c}
  \newcounter{d}
  \foreach \x/\name/\color/\percentage in {36/Fabricated/61a5c2/36\%, 50/Incomplete/89c2d9/50\%, 14/Inconsistency/a9d6e5/14\%}
  {
      \definecolor{color}{HTML}{\color}
      \setcounter{c}{\value{d}}
      \addtocounter{d}{\x}
      \pgfmathsetmacro{\middleangle}{(\thec+\thed)/2}
      \pgfmathsetmacro{\labelradius}{\InsideTextRadius*0.8}
      \pgfmathsetmacro{\percentageradius}{\MiddleTextRadius}
      \coordinate (target_2) at ({\radius*cos(\thec/100*360)},{\radius*sin(\thec/100*360)});
      \coordinate (label_pos2) at ({\labelradius*cos(\middleangle/100*360)}, {\labelradius*sin(\middleangle/100*360)});
      \coordinate (percentage_pos2) at ({\labelradius*cos(\middleangle/100*360)}, {\labelradius*sin(\middleangle/100*360)-0.5});
      \filldraw[line width=\BorderWidth mm, draw=border, fill opacity=1, fill=color] (0,0) -- (target_2) arc ({\thec/100*360}: {\thed/100*360} : \radius cm) -- cycle;

      \node at (label_pos2) {\textbf{\name}};
      \node at (percentage_pos2) {\percentage};
  }
  \end{tikzpicture}}
  \vspace{-20pt}
  \caption*{(b) retrieval heads}
  \label{fig:subfig_b}
\end{minipage}
\caption{Comparison of two sets of error patterns.}
\vspace{-5mm}
\label{fig:error_pattern}
\end{figure}
\paragraph{Findings 2: Error patterns of unfaithfulness induced by masking retrieval heads mirror real error patterns.}
We further manually analyzed the error types triggered by masking retrieval heads and compared them to unfaithfulness error types generated from the model itself (detailed illustration of different error types available in Appendix~\ref{appendix:error_type}). As shown in Figure~\ref{fig:error_pattern}, the results show a notable similarity between these two sets of error patterns, with incomplete hallucinations being the most common, followed by fabricated hallucinations. This similarity suggests that retrieval heads potentially explain contextual faithfulness in LFQA.
\section{Methodology}
\begin{figure*}[h]
    \centering
    \includegraphics[width=1.0\textwidth]{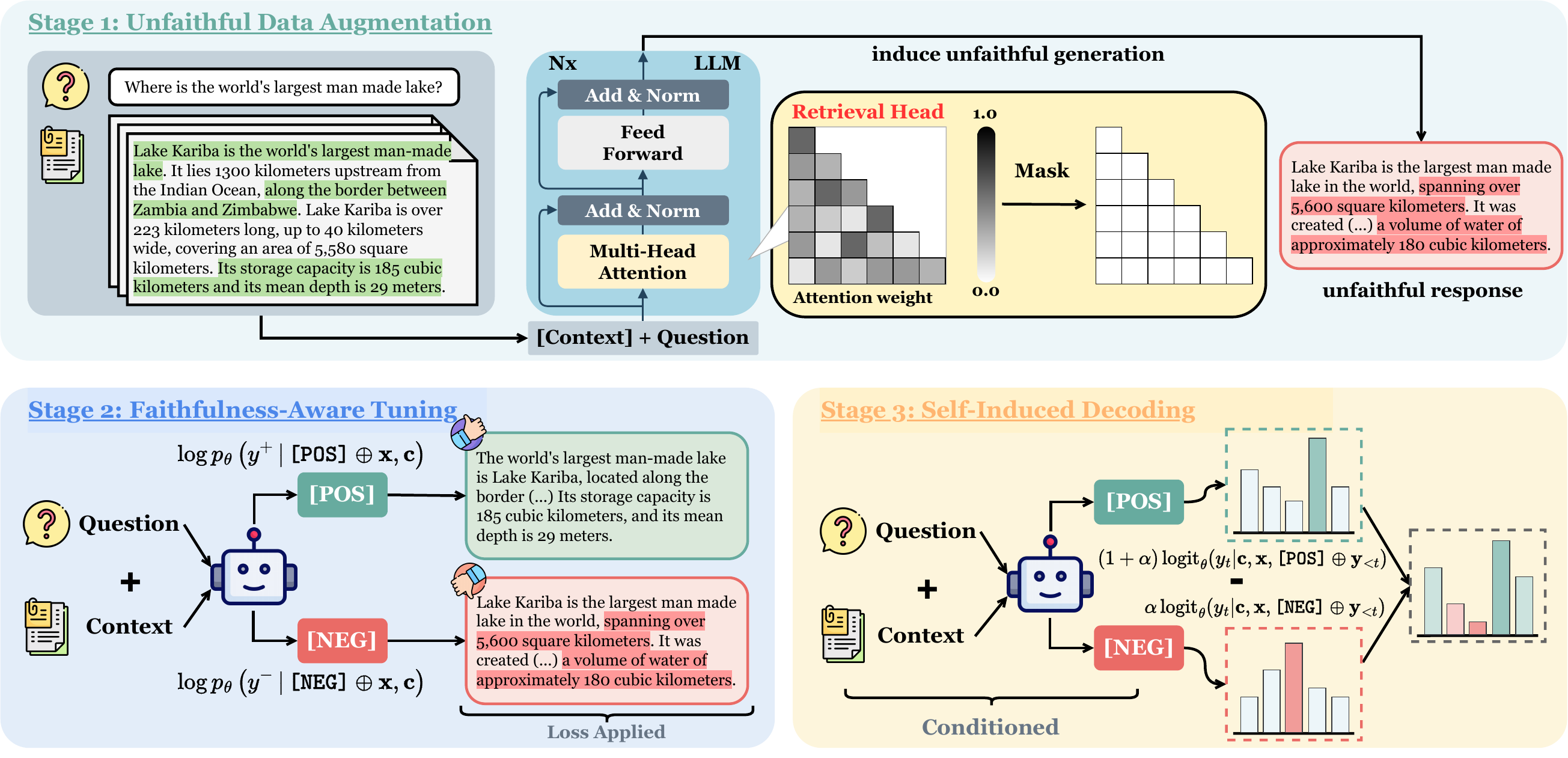}
    \caption{An overview of \ours: (1) unfaithful data augmentation (\S\ref{ssec:negative_augmentation}), which augments unfaithful output by masking out attention heads responsible for contextual faithfulness; (2) faithfulness-aware tuning (\S\ref{ssec:learning_to_discriminate}), which teaches LLMs to explicitly discriminate between faithful and unfaithful outputs; (3) self-induced decoding (\S\ref{ssec:unfaithful_induced_decoding}), which further enhances faithfulness by amplifying the differences between induced contrastive outputs.}
    \label{fig:training-framework}
\end{figure*}
In this section, we introduce \ours, a framework that enhances the contextual faithfulness of LLMs by explicitly teaching them to discriminate between faithful and unfaithful outputs, as illustrated in Figure~\ref{fig:training-framework}. To achieve this, our approach involves augmenting realistic unfaithful samples, which enables the model to learn to distinguish faithful and unfaithful outputs conditioned on different control tokens. These control tokens are further leveraged to induce contrastive outputs and amplify the difference between them via contrastive decoding.
\subsection{Unfaithful Data Augmentation}
\label{ssec:negative_augmentation}
When it comes to negative data augmentation, the key challenge is how to augment samples that simulate the actual error types from the LLM itself.
Prior approaches \citep{mishra2024fine} typically focus on entity-centric permutations, which can result in incoherence and low error coverage, thus limiting their utility in mimicking actual error types.
Inspired by our pilot study in Section (\S\ref{ssec:retrieval_head}), we discovered a salient correlation between LFQA faithfulness and retrieval heads. This finding motivates us to generate more realistic unfaithful outputs by simply masking out retrieval heads of LLMs.

Formally, we define the attention matrix for a given layer as $\mathbf{A} \in \mathbb{R}^{H \times L \times L}$, where $H$ is the number of attention heads, and $L$ is the sequence length. 
Let $\mathcal{R}$ represent the indices of the top-N retrieval heads identified by the detection algorithm in \citet{wu2024retrieval}.
The masking operation is defined as:
\[
\mathbf{A}'_{hij} = 
\begin{cases} 
0 & \text{if } h \in \mathcal{R} \\
\mathbf{A}_{hij} & \text{otherwise}
\end{cases}
\]
where $\mathbf{A}'$ is the modified attention matrix with weights of the top-N retrieval heads set to zero. $h$ indexes the attention heads. Here, we set $N=100$.

\subsection{Faithfulness-Aware Tuning}
\label{ssec:learning_to_discriminate}
Beyond simply fine-tuning LLMs on faithful data, which encourages them to imitate positive behaviors, we argue that the key to improving faithfulness lies in enhancing the model's ability to perceive and distinguish different types of unfaithfulness errors.
To this end, we propose \textbf{F}aithfulness-\textbf{A}ware \textbf{T}uning (\textbf{FAT}), which aims at teaching LLMs to discriminate faithful responses from unfaithful ones.

To explicitly distinguish the contrastive outputs, we draw inspiration from controllable text generation \citep{lu2022quark} and introduce two special \textit{control codes}, \posi and \nega.
These special control tokens signal the generation of either faithful or unfaithful output.
Specifically, we prepend a prefix at the beginning of the generation, \eg, \posi for the faithful output $\yPos$ and \nega for the unfaithful output $\yNeg$ and then supervise the model to generate corresponding outputs conditioned on the control codes.
Our training objectives are twofold:
\begin{align}\label{eq:sft}
    & \mathcal{L}(\theta) = -\underbrace{\mathbb{E}_{(\mathbf{x}, \mathbf{c}, \yPos)}
    \left[
      \log p_\theta \left( \yPos \,|\, \text{\posi} \oplus \mathbf{x}, \mathbf{c} \right)
    \right]}_\text{enhancing faithfulness perception} \notag \\
    & - \underbrace{\mathbb{E}_{(\mathbf{x}, \mathbf{c}, \yNeg)}
    \left[
      \log p_\theta \left( \yNeg \,|\, \text{\nega} \oplus \mathbf{x}, \mathbf{c} \right)
    \right]}_\text{enhancing unfaithfulness perception}
\end{align}
where $\mathbf{x}$ and $\mathbf{c}$ represent the question and context, respectively, and $\oplus$ denotes the concatenation of the control code with the question and context.
\subsection{Self-Induced Decoding}
\label{ssec:unfaithful_induced_decoding}
During the inference phase, these trained control tokens enable the model to tailor its behavior.
When prepending \posi, it steers the model to generate faithful responses. Conversely, it can also induce a hallucinated unfaithful response when prepending \nega.
Inspired by studies on contrastive decoding \citep{li2023contrastive, shi2024trusting}, we introduce self-induced decoding (\textbf{SID}), which aims to further enhance faithfulness by amplifying the differences between induced contrastive outputs.

Concretely, consider an LLM parameterized by $\theta$, the model takes a question $\mathbf{x}$ and source documents $\mathbf{c}$ as input to generate a long-form response $\mathbf{y}$. 
We prepend the two control codes before generation, the next-token probability distribution is determined by amplifying the predictions from the faithful prediction and downplaying the induced unfaithful one, which can be formed as:
\begin{align}\label{eq:contrastive_decoding}
    y_t \sim & \text{Softmax} \big[ \big( (1+\alpha) \, \text{logit}_{\theta}(y_t | \mathbf{c}, \mathbf{x}, \footnotesize\text{\posi} \oplus \mathbf{y}_{<t}) \notag \\ 
    & - \alpha \,\text{logit}_{\theta}(y_t | \mathbf{c}, \mathbf{x}, \footnotesize\text{\nega} \oplus \mathbf{y}_{<t}) \big) / \tau \big]
\end{align}
Here, $\tau$ is the temperature in the sampling decoding strategy.
A larger $\alpha$ indicates more weight focus on the context $\mathbf{c}$ and $\alpha=0$ reduces to vanilla decoding. We set $\alpha=0.2$ in our experiments.

\section{Benchmark: \benchmark}
\label{sec:ground_bench}
In this section, we introduce \benchmark, a benchmark tailored for evaluating the faithfulness of LFQA, and present the data collection pipeline as well as the manual effort for constructing \benchmark.

\subsection{Data Collection}
Several datasets~\citep{fan2019eli5, stelmakh2022asqa} have been introduced to assess the performance of LLMs on retrieval-augmented LFQA across various domains and evaluation dimensions. However, no existing dataset is specifically designed to measure faithfulness. This gap largely stems from the lack of guarantee that the retrieved documents contain sufficient information to answer the questions, a primary cause of unfaithfulness in LFQA \citep{chen2023understanding}. This limitation complicates the evaluation of LLMs' faithfulness.

To bridge the gap, we introduce \benchmark, an aggregated benchmark composed of five LFQA datasets, including ELI5-WebGPT~\citep{nakano2021webgpt}, ExpertQA~\citep{malaviya2023expertqa}, HAGRID~\citep{kamalloo2023hagrid}, CLAPNQ~\citep{rosenthal2024clapnq}, and QuoteSum~\citep{schuster2024semqa}. 
These datasets feature a wide range of queries, retrieval sources, and varying levels of difficulty, providing a comprehensive evaluation testbed.
For detailed statistics, see Appendix~\ref{appendix:datasets}.

\subsection{Dataset Specific Settings}
To facilitate the evaluation of faithfulness in LFQA, we implement a controlled setting for each dataset to ensure the provided documents contain sufficient information to answer the questions.

\paragraph{ELI5-WebGPT} consists of questions sourced from the “Explain Like I’m Five” subreddit. We use golden documents collected by human annotators via commercial search engines, which are deemed relevant and sufficiently informative to answer the questions. Each question is also equipped with human-labeled answers.

\paragraph{ExpertQA} contains information-seeking questions formulated by experts spanning 32 fields, each paired with relevant documents and expert-verified answers. Considering that some of the annotated data have missing textual forms of evidence, we manually curated data with expert-revised evidence for \benchmark.

\paragraph{HAGRID} includes questions designed for information-seeking scenarios, each accompanied by a set of manually labeled relevant documents and an answer generated by LLMs. We manually select those entries the answers are considered both informative and well-grounded by human evaluators for \benchmark.

\paragraph{CLAPNQ} is built on real web search queries sampled from Natural Questions \citep{kwiatkowski2019natural}. It features a gold document for each question with a corresponding grounded long-form answer. Additionally, we retrieve four additional documents from Wikipedia to create a multi-source information synthesis scenario.

\paragraph{QuoteSum} is a semi-extractive LFQA dataset that involves questions with relevant documents and human-written answers derived explicitly from extracted spans across multiple sources. Specifically, we include the test subset in \benchmark.

\section{Experimental Setup}
\label{sec:experiments}

\subsection{Evaluation}
\definecolor{steelbluev2}{HTML}{DAE8FC}
\definecolor{steelblue}{HTML}{82B0D2}
\definecolor{mygray}{HTML}{808080}

\begin{table*}[h]
\centering
\resizebox{\textwidth}{!}{
\begin{tabular}{lccccccccccc}
\toprule[1.0pt]
\multicolumn{1}{l}{\multirow{2}{*}{Method}} & \multicolumn{2}{c}{\textbf{CLAPNQ}} & \multicolumn{2}{c}{\textbf{ExpertQA}} & \multicolumn{2}{c}{\textbf{HAGRID}} & \multicolumn{2}{c}{\textbf{ELI5-WebGPT}} & \multicolumn{2}{c}{\textbf{QuoteSum}} & \multirow{2}{*}{\textbf{Avg. Faith.}} \\ \cmidrule(l){2-3} \cmidrule(l){4-5} \cmidrule(l){6-7} \cmidrule(l){8-9} \cmidrule(l){10-11} 
\multicolumn{1}{c}{} & RL. & Faith. & RL. & Faith. & RL. & Faith. & Claim. & Faith. & SEM. & Faith.  \\ \midrule
   GPT-4o & 40.53  & 91.81 & \textcolor{mygray}{\textbf{46.34}} & 69.48 & \textcolor{mygray}{\textbf{57.76}} & \textcolor{mygray}{\textbf{90.86}}   & \textcolor{mygray}{\textbf{59.04}}  & 81.00 & 42.56 & 78.51 & 82.33 \\
   GPT-4o-mini & 37.72 & 90.35 & 45.30 & 66.53 & 54.87 & 87.94 & 56.09 & 81.89 & 40.74 & 78.16 & 80.97\\ 
   \midrule
   Llama-3.1-70B-Instruct & \ul{39.44} & 88.64 & \textbf{43.02} & 69.35 & 49.21 & 79.08 & \ul{51.66} & 74.87 & 41.24 & 67.42 & 75.87\\ 
   Mistral-NeMo-12B-Instruct & 35.28 & 78.71 & 42.76 & 54.19 & \ul{53.05} & 80.16 & \textbf{53.84} & 65.06 & 39.50 & 69.85 & 69.59 \\
   Llama-3.1-8B-Instruct & 17.14 & 58.47 & 31.67 & 51.22  & 16.47 & 55.80 & 47.11 & 55.74 & 25.96 & 41.70 & 52.59\\ 
   \midrule
\textbf{\textit{Llama-2-7B}} \\
\ \ \ \ \ + SFT & 37.43 & 82.46  & 42.47 & 68.14  & 50.89 & 78.14 & 47.85 & 73.25 & 40.81 & 62.93  & 72.98\\
\ \ \ \ \ + RECOMP & 30.88 & 63.52  & 39.72  & 51.95 & 38.30 & 62.72 & 40.22 & 49.45 & 37.61 & 54.98 & 56.52\\
\ \ \ \ \ + Self-RAG & 36.22  & 86.11  & 26.31 & 58.53 & 43.14 & 71.77 & 19.19  & 63.22 & 40.64 & 63.36  & 68.60\\ 
\rowcolor{steelblue!33}
\ \ \ \ \ + \ours & 38.71 & \ul{90.34}  & 39.34  & \ul{75.49} & 49.59 & \ul{86.43} & 42.07 & \ul{82.51} & \ul{42.81} & \ul{76.98} & \ul{82.35}  \\ \midrule
\textbf{\textit{Llama-2-13B}}   \\
\ \ \ \ \ + SFT  & 37.18 & 84.72  & \ul{42.85}  & 70.87 & \textbf{53.88} & 78.07 & 47.72 & 71.61 & 40.64 & 66.72 & 74.40 \\
\ \ \ \ \ + RECOMP & 30.62 & 64.91  & 40.18 & 52.18  & 37.65 & 59.85 & 39.36 & 51.96 & 37.62 & 51.75 & 56.13\\
\ \ \ \ \ + Self-RAG & 31.34 & 61.62  & 21.72  & 56.05 & 42.62 & 56.27 & 10.82 & 36.87 & 40.69 & 67.40 & 55.64 \\ \rowcolor{steelblue!33}
\ \ \ \ \ + \ours        & \textbf{41.14} & \textbf{92.14}  & 39.65  & \textbf{77.86} & 50.64 & \textbf{87.18} & 42.84 & \textbf{82.94} & \textbf{42.82} & \textbf{78.73} & \textbf{83.77} \\
\bottomrule[1.0pt]
\end{tabular}
}
\caption{Experimental results on \benchmark. \textbf{Bold} and \ul{underline} numbers indicate the best performance and second performance among non-proprietary models. And \textcolor{mygray}{\textbf{gray-colored}} bold text indicates the best proprietary model when they outperforms all non-proprietary models.}
\label{tab:main_results}
\end{table*}

All methods are evaluated on \benchmark, combining both automatic and human evaluation.

\subsubsection{Automatic Metrics}
We evaluate long-form responses mainly on two dimensions: \textbf{Overall Quality} and \textbf{Faithfulness}.

\paragraph{Overall Quality} measures the alignment of the model response with the human-labeled response. For CLAPNQ, ExpertQA, and HAGRID, we employ ROUGE-L to evaluate overall quality. For ELI5-WebGPT, we calculate the claim recall between the model response and golden sub-claims, following \citet{chen2023understanding}. Regarding QuoteSum, we employ SEMQA \citep{schuster2024semqa} to measure the overall quality. For further details about the automatic metrics, please refer to Appendix~\ref{appendix:automatic_metrics}.

\paragraph{Faithfulness} measures the extent to which the model response is grounded in the provided context. We use MiniCheck \citep{tang2024minicheck}\footnote{\url{https://huggingface.co/bespokelabs/Bespoke-MiniCheck-7B}} to assess the consistency between each statement in the response and the documents. The average for all statements in each dataset is reported as the FaithScore.

\subsubsection{Human Evaluation.}
Recent research \citep{xu2023a} highlighted the challenges of evaluating LFQA. To this end, we engage human annotators to manually analyze 200 generations from selected methods across the following dimensions.
Each generation is rated on a 5-point Likert scale for each dimension. For more details, please refer to Appendix~\ref{appendix:human_evaluation}.

\paragraph{Faithfulness} evaluates whether the answer is \textit{fully supported}, \textit{partially supported}, or \textit{not supported} by the provided documents.
\paragraph{Completeness} measures whether all relevant information in the context is included to generate an informative answer to the question.

\subsection{Baselines}
We compare \ours with the following baselines.

\paragraph{Prompting-based Method} simply prompts LLMs to generate long-form responses that are faithful to the provided context for each question. Specifically, we evaluate several SOTA proprietary LLMs with carefully designed prompts, including GPT-4o and GPT4o mini\footnote{\texttt{Specifically, we utilize gpt-4o-2024-08-06} and \texttt{gpt-4o-mini-2024-07-18} version for evaluation.}, as well as open-source models of varying sizes, covering models from the Llama-3.1 family and Mistral-NeMo-12B-Instruct.

\paragraph{Supervised Fine-Tuning (SFT)} directly fine-tunes the model with high-quality LFQA data, aiming at teaching the model to utilize contextual information to generate faithful responses.

\paragraph{RECOMP} \citep{xu2023recomp} train an abstractive summarization model to filter irrelevant information in the context, avoiding the impact of noisy documents on faithful generation.

\paragraph{Self-RAG} \citep{asai2024selfrag} utilize special reflection tokens to teach the model to assess the retrieval quality and self-reflect with its generation, thereby improving faithfulness.

\subsection{Implementation Details}
To ensure a fair comparison, we employ the \texttt{Llama-2-7B} and \texttt{Llama-2-13B} as backbones for all the training-based baselines. 
We train models using the long-form split of FRONT dataset \citep{huang2024learningfinegrainedgroundedcitations}, which consists of diverse real-world user queries for information-seeking. 
In the decoding stage, to avoid low-quality outputs during long-form generation, we uniformly adopt a sampling decoding strategy with a temperature of 1 and top-p of 0.95.
For more implementation details of baselines and \ours, please refer to Appendix~\ref{appendix:implementation}.
\section{Results}
\subsection{Main Results}
\definecolor{steelbluev2}{HTML}{DAE8FC}
\definecolor{steelblue}{HTML}{82B0D2}

\begin{table*}[h]
\centering
\resizebox{\textwidth}{!}{%
\begin{tabular}{lccccccccccc}
\toprule[1.0pt]
\multicolumn{1}{l}{\multirow{2}{*}{Method}} & \multicolumn{2}{c}{\textbf{CLAPNQ}} & \multicolumn{2}{c}{\textbf{ExpertQA}} & \multicolumn{2}{c}{\textbf{HAGRID}} & \multicolumn{2}{c}{\textbf{ELI5-WebGPT}} & \multicolumn{2}{c}{\textbf{QuoteSum}} & \multirow{2}{*}{\textbf{Avg. Faith.}} \\ \cmidrule(l){2-3} \cmidrule(l){4-5} \cmidrule(l){6-7} \cmidrule(l){8-9} \cmidrule(l){10-11} 
\multicolumn{1}{c}{} & RL. & Faith. & RL. & Faith. & RL. & Faith. & Claim. & Faith. & SEM. & Faith.  \\ \midrule
\textbf{\textit{Llama-2-7B}} \\
\rowcolor{gray!10}
\ \ \ \ \ + \ours & \textbf{38.71} & \textbf{90.34}  & 39.34  & \textbf{75.49} & 49.59 & \textbf{86.43} & 42.07 & \textbf{82.51} & \textbf{42.81} & \textbf{76.98} & \textbf{82.35}  \\
\ \ \ \ \ \ \ w/o SID & 38.03 & 88.57  & 41.17  & 73.37 & 48.96 & 84.54 & 43.05 & 79.00 & 41.81 & 74.68 & 80.03  \\
\ \ \ \ \ \ \ w/o FAT & 37.43 & 82.46  & \textbf{42.47}  & 68.14 & \textbf{50.89} & 78.14 & \textbf{47.85} & 73.25 & 40.81 & 62.93 & 72.98  \\ \midrule
\textbf{\textit{Llama-2-13B}}   \\
\rowcolor{gray!10}
\ \ \ \ \ + \ours & \textbf{41.14} & \textbf{92.14}  & 39.65  & \textbf{77.86} & 50.64 & \textbf{87.18} & 42.84 & \textbf{82.94} & \textbf{42.82} & \textbf{78.73} & \textbf{83.77} \\
\ \ \ \ \ \ w/o SID        & 37.74 & 88.33  & 40.28  & 73.23 & 49.55 & 85.40 & \textbf{48.13} & 79.54 & 42.43 & 75.58 & 80.42 \\
\ \ \ \ \ \ w/o FAT        & 37.18 & 84.72  & \textbf{42.85}  & 70.87 & \textbf{53.88} & 78.07 & 47.72 & 71.61 & 40.64 & 66.72 & 74.40 \\
\bottomrule[1.0pt]
\end{tabular}
}
\caption{Ablation Experimental results on \benchmark. The best results for each dataset are in \textbf{bold}.}
\label{tab:ablation_results}
\end{table*}

We present the results of \benchmark in Table~\ref{tab:main_results}. 
\paragraph{Exisiting LLMs are struggling to generate faithful responses.} We can observe that GPT-4o achieves the best performance across five datasets, with an average faithfulness score of 82.33\%, while GPT-4o-mini ranks in the second tier. The performance of open-source models shows a clear downward trend as the model size decreases. Even for the most powerful LLMs, achieving full contextual faithfulness remains a significant challenge. Crucially, when it comes to the smaller LLM, \eg, Llama-3.1-8B-Instruct, it exhibits a substantial performance gap of 23.28\% compared to its larger counterpart. This indicates the challenge for smaller LMs to maintain faithfulness to the context.

\paragraph{LLMs fine-tuned to utilize contextual information achieve better faithfulness.}
Compared with other training-based baselines, simply fine-tuning LLMs to utilize contextual information (SFT) results in more faithful long-form responses, leading to an average improvement of 30.84\% and 20.05\% compared with RECOMP and Self-RAG, respectively.
This finding highlights the effectiveness of training models with attribution in mind to enhance faithfulness, which also aligns with prior work \citep{chen2023understanding}.

\paragraph{\ours significantly improves contextual faithfulness, even surpassing GPT-4o.} 
For both the 7B and 13B models, \ours significantly improves faithfulness compared to all training-based baselines in \benchmark, leading to an average improvement of 12.84\% and 12.59\% in faithfulness for the 7B and 13B models, respectively. Notably, \ours even outperforms the state-of-the-art GPT-4o by 1.74\%.
This indicates that teaching the model to distinguish between faithfulness and unfaithfulness can effectively establish a mechanism for faithful generation.

\subsection{Ablation Studies}
We conduct ablation studies to verify the effectiveness of \ours, and results are shown in Table~\ref{tab:ablation_results}.
\paragraph{Effect of faithfulness-aware tuning.} 
To understand the significance of faithfulness-aware tuning (FAT), we compare \ours with models that are solely fine-tuned using faithful samples. As shown in Table~\ref{tab:ablation_results}, omitting FAT leads to a substantial decrease in faithfulness (82.35\% $\rightarrow$ 72.98\% in 7B and 83.77\% $\rightarrow$ 74.40\% in 13B, respectively). 
This highlights the effectiveness of unfaithful samples in enhancing the perception of faithfulness in LLMs.

\paragraph{Effect of self-induced decoding.}
To demonstrate the effectiveness of self-induced decoding (SID), we compare \ours with models that use the vanilla decoding strategy ($\alpha=0$). As indicated in Table~\ref{tab:ablation_results}, SID further improves average faithfulness by 2.90\% and 4.17\% in 7B and 13B models, respectively.
Besides, we further investigate the effect of varying $\alpha$ on faithfulness by adjusting $\alpha$ from 0.1 to 0.5. 
As shown in Figure~\ref{fig:ablation_difference_alpha} (a), as $\alpha$ increases, the model's performance on faithfulness first and gradually decreases, reaching its best performance at $\alpha=0.2$.
We further compare SID with context-aware decoding (CAD), which amplifies the difference between output probabilities with and without context. 
As shown in Figure~\ref{fig:ablation_difference_alpha} (b), SID slightly outperforms CAD in faithfulness. 
This indicates that SID, by leveraging more diverse model-intrinsic error types induced by \nega, more effectively improves model faithfulness than merely amplifying context-aware contrasts.
\definecolor{red}{RGB}{172,21,28}
\definecolor{blue}{RGB}{39,89,167}
\definecolor{red1}{RGB}{203,104,104}
\definecolor{blue1}{RGB}{104,155,203}
\definecolor{color1}{RGB}{235,164,122}
\definecolor{color2}{RGB}{78,172,183}

\begin{figure}[t]
    \centering
    \vspace{1.0mm}
\begin{tikzpicture}
    \scriptsize{
    \begin{axis}
    [
        anchor=north west,
        at={(-3.5em,-5em)},
        ymajorgrids,
        xmajorgrids,
        grid style=dashed,
        width=.27\textwidth,
        height=.26\textwidth,
        yticklabel style={/pgf/number format/precision=2,/pgf/number format/fixed zerofill,scale=1.0},
        xmax=2100,
        xmin=300,
        ymin=0.775,
        ymax=0.845,
        xtick={400,800,1200,1600,2000},
        xticklabels={0.1,0.2,0.3,0.4,0.5},
        ytick={0.78,0.79,0.80,0.81,0.82,0.83,0.84},
        xlabel={\scriptsize{(a) Hyperparameter $\alpha$}},
        xlabel style={scale=1.2, yshift=0.8em, xshift=0.1em},
        ylabel=\footnotesize{\scriptsize Avg. FaithScore},
        ylabel style={yshift=-1.9em, scale=1.2},
legend style={at={(0.83,0.37)}, anchor=north east, font={\tiny}, cells={anchor=west}, fill opacity=0.8, scale=1.0}
        ]

\addplot[red,mark=pentagon*,,mark size=1.8pt,thick,mark options={fill=white,draw=red,line width=1pt}] coordinates {(400,0.8115) (800,0.8235) (1200,0.8103) (1600,0.8068) (2000,0.7993)};
        \addlegendentry{\scalebox{1.2}{\ours-7B}}

        \addplot[blue,mark=*,mark size=1.8pt,thick,mark options={fill=white,draw=blue,line width=1pt}] coordinates {(400,0.8296) (800,0.8377) (1200,0.8301) (1600,0.8274) (2000,0.8274)};
        \addlegendentry{\scalebox{1.2}{\ours-13B}}   
    \end{axis}
	
  \begin{axis}[
    at={(11.5em,-15.5em)},
    anchor=south west,
    ymajorgrids,
    grid style=dashed,
    legend style={at={(0.02,0.65)}, anchor=south west},
    legend cell align={left},
    ybar,
    enlarge x limits=0.5,
    xtick align=inside,
    height=.26\textwidth,
    width=.27\textwidth,
    bar width=1.3em,
    xlabel={\scriptsize{(b) Decoding Strategy}},
    xlabel style={scale=1.2, yshift=0.8em, xshift=0.1em},
    ylabel=\footnotesize{\scriptsize Avg. FaithScore},
    ylabel style={scale=1.2, yshift=0.5em},
    symbolic x coords={{1}, {2}},
    xtick=data,
    ymin=0.795,
    ymax=0.865,
    ytick={0.80, 0.81, 0.82, 0.83,0.84, 0.85, 0.86},
    nodes near coords align={vertical},
    xticklabels={7B, 13B},
    ylabel style={yshift=-2em},
    yticklabel style={/pgf/number format/fixed,/pgf/number format/fixed zerofill,/pgf/number format/precision=2,rotate=0,scale=1.0},
    legend style={yshift=0.2em,xshift=4.2em,font={\tiny},cells={anchor=west},fill opacity=0.8, scale=1.0, legend columns=1}
    ]
    \addplot[fill=color1, draw=color1, area legend] coordinates {({1},0.8235) ({2},0.8377)};
    \addlegendentry{\scalebox{1.0}{\textsc{HID}.}}
    \addplot[fill=color2, draw=color2, area legend] coordinates {({1},0.81) ({2},0.828)};
    \addlegendentry{\scalebox{1.0}{\textsc{CAD}.}}
  \end{axis}
}   
\end{tikzpicture}
    \vspace{-3mm}
    \caption{
    Ablation study on hyperparameter $\alpha$ and the decoding strategy in self-induced decoding.
    }
    \label{fig:ablation_difference_alpha}
\end{figure}
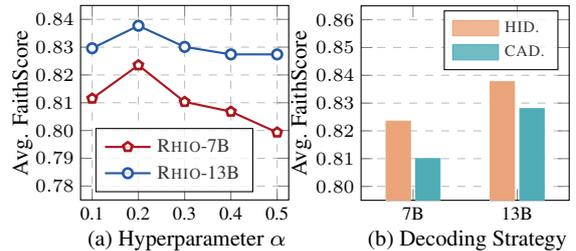
\subsection{Further Analysis}
\paragraph{Does masking out retrieval heads outperform other augmentation strategies?} 
\definecolor{red1}{RGB}{203,104,104}
\definecolor{blue1}{RGB}{104,155,203}
\definecolor{red2}{RGB}{236,121,122}
\definecolor{blue2}{RGB}{108,179,211}
\definecolor{uorange}{RGB}{247,175,89}
\definecolor{upurple}{RGB}{148,137,250}
\definecolor{pink1}{HTML}{FFF0F6}
\definecolor{pink2}{HTML}{FCC2D7}
\definecolor{pink3}{HTML}{FAA2CA}

\begin{figure}[!t]
\centering
\begin{tikzpicture}
\scriptsize{
\begin{axis}[
    ymajorgrids,
    xmajorgrids,
    grid style=dashed,
    ylabel={\footnotesize{Avg. FaithScore (\%)}},
    legend style={at={(0.5,1.03)}, anchor=south, legend columns=-1, nodes={scale=0.7, transform shape},
        },
    ybar,
    enlarge x limits=0.6,
    xtick align=inside,
    height=.3\textwidth,
    width=.5\textwidth,
    bar width=2.0em,
    nodes near coords,
    nodes near coords align={vertical},
    nodes near coords style={font=\tiny, scale=0.8,/pgf/number format/fixed, /pgf/number format/precision=1},
    every node near coord/.append style={/pgf/number format/.cd, fixed, fixed zerofill, precision=1},
    xlabel={\footnotesize{Model Sizes}},
    symbolic x coords={0,1},
    xtick=data,
    ymin=69.0,
    ymax=83.0,
    ytick={70.0, 72.0, 74.0, 76.0, 78.0, 80.0, 82.0},
    yticklabels={70.0, 72.0, 74.0, 76.0, 78.0, 80.0, 82.0},
    xticklabels={7B, 13B},
    ylabel style={yshift=-2.em},
    xlabel style={yshift=1em,align=center},
    yticklabel style={/pgf/number format/fixed},
]
    \addplot[fill=red1,draw=red1, area legend] coordinates {(0,73.4) (1,75.2)};
    \addlegendentry{Entity.}
    \addplot[fill=red2, draw=red2, area legend] coordinates {(0,73.5) (1,75.4)};
    \addlegendentry{Relation.}
    \addplot[fill=blue1,draw=blue1, area legend] coordinates {(0,76.4) (1,77.5)};
    \addlegendentry{Prompt.}
    \addplot[fill=blue2,draw=blue2, area legend] coordinates {(0,80.03) (1,80.42)};
    \addlegendentry{Retrieval Head}
\end{axis}
}
\end{tikzpicture}
 \vspace{-3mm}
\caption{Ablation study on different negative sample augmentation strategies.}
\label{fig:ablation_negative_sample}
\vspace{-5mm}
\end{figure}
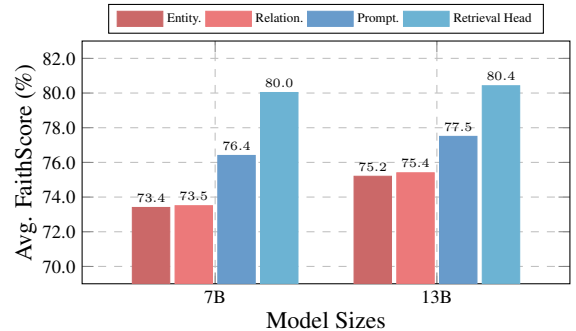
To demonstrate the effectiveness of utilizing negative samples triggered by masking out retrieval heads in improving faithfulness, we conduct a comparative ablation study by employing various negative sample augmentation strategies, including entity replacement, relation corruption, and direct prompting.
Detailed implements for these augmentation strategies are presented in Appendix~\ref{appendix:negative_sample}.
As shown in Figure \ref{fig:ablation_negative_sample}, entity and relationship perturbations yield only marginal improvement. Meanwhile, compared to prompting methods, unfaithful samples induced by retrieval heads lead to more significant enhancements in faithfulness.

\paragraph{Does self-induced unfaithful samples perform better?}
To analyze this, we separately trained two models using unfaithful outputs generated by masking retrieval heads from different models. As shown in Figure~\ref{fig:ablation_self_induced}, self-induced negative samples lead to more improvement in faithfulness. We posit that self-induced samples, embedded with intrinsic model-specific errors, provide a more effective learning environment for the models to distinguish between faithful and unfaithful outputs.
In addition, we explore the impact of the number of masked retrieval heads on performance in Appendix~\ref{appendix:analysis}.

\section{Human Evaluation}
\begin{table}[t]
\small
\centering
\resizebox{\linewidth}{!}{
\begin{tabular}{@{}lcccc@{}}
\toprule
& \multicolumn{3}{c}{Faithfulness} & \multicolumn{1}{c}{Completeness} \\
\cmidrule(lr){2-4} \cmidrule(lr){5-5}
& Full & Partial & No & Rate\\
\midrule
GPT-4o   & \ul{86.5\%} & \ul{8.1\%} & \ul{5.4\%}   & \textbf{4.2}   \\
Llama-3.1-70B-Instruct & 82.6\% & 10.2\% & 7.2\% & 3.7 \\
SFT-13B & 76.8\% & 13.8\% & 9.4\% & 3.2 \\
\ours-13B (Ours)    & \textbf{87.5\%}  & \textbf{7.6\%} & \textbf{4.9\%} & \ul{3.8} \\
\bottomrule
\end{tabular}
}
\caption{Human evaluation results on faithfulness and completeness. \textbf{Bold} numbers indicate the best performance. ``$\_$'' indicates the second-best performance.}
\vspace{-4mm}
\label{tab:human_evaluation}
\end{table}

Human evaluation results, detailed in Table~\ref{tab:human_evaluation}, indicate that \ours generates significantly more grounded responses compared to all baselines.
We show some case studies of \ours in Appendix~\ref{appendix:case_study}.

\section{Related Work}
Recently, the demand for contextual LLMs has continued to grow, particularly in retrieval-augmented generation (RAG) scenarios. Despite its significant progress, hallucinations~\citep{huang2023a} remains a critical challenge.
Generally, hallucination in LLMs can be categorized into two types: \textit{factuality hallucination}, where generated content deviates from established world knowledge, and \textit{faithfulness hallucination}, where the generated response is inconsistent with the provided context.
In this work, we focus on faithfulness hallucination, particularly in long-form question-answering (LFQA)~\citep{fan2019eli5, han2024ragqaarenaevaluatingdomain}.

Unlike factoid QA~\citep{kwiatkowski2019natural, mallen2023when}, where the answer units often appear as short-form entities, LFQA is a challenging task, which requires synthesizing relevant information from retrieved documents to produce a complex, paragraph-length answer.
Contextual faithfulness~\citep{nguyen2024sfrragcontextuallyfaithfulllms} in LFQA measures whether the model generation is supported by the provided context~\citep{huang2024advancing}. Recent study~\citep{chen2023understanding} reveals that unfaithfulness in LFQA mainly comes from retrieval failure, hallucinated facts, and incorrect synthesis. Moreover, \citet{alessandro2024groundedness} found that the unfaithfulness phenomenon is prevalent across different model sizes and highlighted the propensity of LLMs to blend correct information with hallucinated content. 

Numerous effects have been made to improve contextual faithfulness in LFQA. 
RECOMP~\citep{xu2023recomp} leveraged a summarization model to filter out the irrelevant context in retrieved documents before the generation process.
Self-RAG~\citep{asai2024selfrag} designed special reflection tokens to teach models self-reflection abilities. This enables the model to adaptively retrieve documents on-demand, and criticize its own output to improve the faithfulness and overall quality.
\citet{shi2024trusting, wang2024adacadadaptivelydecodingbalance} combine pointwise mutual information with contrastive decoding~\citep{li2023contrastive} to let the model focus more on contextual information to reduce unfaithfulness. Despite these efforts, these approaches are more of a compensatory way to improve faithfulness, rather than developing models with a \textit{built-in} mechanism for faithfulness generation.

In contrast to these methods, our work draws inspiration from retrieval heads \citep{wu2024retrieval}, which play a fundamental role in contextual faithfulness. This allows us to augment realistic unfaithful samples and improve the perception of faithfulness by teaching the model to learn to distinguish between faithful and unfaithful responses.
\section{Conclusion}
This work proposes \ours, a framework designed to enhance the contextual faithfulness of LLMs.
\ours first augments realistic unfaithful outputs by selectively masking out attention heads responsible for information retrieval.
Given both faithful and unfaithful samples, \ours trains the model to explicitly distinguish between them, conditioned on faithfulness control codes.
\ours further enhances faithfulness with contrastive decoding, aimed at amplifying the differences between contrastive outputs induced by control codes.
Additionally, we also introduce \benchmark, a comprehensive benchmark compiled from five diverse LFQA datasets, providing a controllable evaluation of faithfulness in LLMs.
Extensive evaluations on \benchmark demonstrate that \ours significantly improves faithfulness, even surpassing GPT-4o.

\section*{Limitations}
This work exhibits several limitations worth noting. 
Firstly, to ensure a fair comparision, our experiments primarily employ the Llama-2 series models, which are utilized in the baseline models, \eg, Self-RAG and RECOMP. Future work will involve extending our method to encompass additional series models.
Secondly, to facilitate the evaluation of faithfulness in retrieval-augmented LFQA, \benchmark is constructed within a controlled setting, where the context is assured to contain sufficient information to answer the question. Consequently, faithfulness hallucination caused by retrieval failures is not discussed in \benchmark. We aim to extend the benchmark to cover a diverse range of task scenarios in future work, thus providing a more comprehensive evaluation of contextual faithfulness in LLMs.
Thirdly, while our findings demonstrate that retrieval heads can produce a diverse array of realistic and model-intrinsic unfaithful errors, the current study does not control the generation of specific types of unfaithful errors. Future research could explore augmenting specific error types by delving into the distinct contributions of various retrieval heads.

\section*{Ethics Statement}
In this work, all data used for constructing \benchmark derive from open-source public datasets, and no additional collection of sensitive information was conducted. Throughout the experimental process, all data and models were strictly utilized following their intended purposes and respective licenses. Our methodology aims to enhance the contextual faithfulness of retrieval-augmented LLMs, which positively impacts information-seeking systems by improving the transparency of LLMs in real-world applications. However, when deployed, our approach still carries inherent issues associated with LLMs, such as the potential for generating biased, harmful, or offensive output. Aside from this, to the best of our knowledge, there are no additional ethical issues associated with this paper.

\bibliography{custom}

\appendix

\section{Preliminary Study of Retrieval Heads}
\label{appendix:retrieval_head}

\subsection{Experiments for Randomly Masking}
\label{appendix:random_mask}

We provide the experimental results of random masking in Tabel~\ref{tab:random_mask}. Specifically, we randomly mask out top-100 non-retrieval heads for each model. We can observe that randomly masking out non-retrieval heads does not have a significant impact on the faithfulness of the model.

\begin{table}[!htbp]
\small
\centering
\resizebox{\linewidth}{!}{
\begin{tabular}{@{}lccc@{}}
\toprule
\textbf{Model} & \textbf{\# R.(0)} & \textbf{\# R.(100)} & \textbf{\# Non-R.(100)}\\ 
\midrule
Llama-2-7B-Chat & 80.14 & 35.85 & 77.12  \\
Llama-2-13B-Chat & 80.09  & 41.09 & 75.13  \\
Llama-2-70B-Chat & 86.36 & 51.49 & 88.16  \\
\bottomrule
\end{tabular}
}
\caption{Experimental results of randomly masking. \textbf{R.(0)} represents masking 0 retrieval heads, \textbf{R.(100)} represents masking the top-100 retrieval heads, while \textbf{Non-R.(100)} represents randomly masking 100 non-retrieval heads. The values in the table represent the average faithfulness of the model response.}
\vspace{-4mm}
\label{tab:random_mask}
\end{table}

\subsection{Analysis of Error Patterns}
\label{appendix:error_type}
To analyze the relationship between error types caused by masking retrieval heads and error types of unfaithful responses generated by the model itself, we extracted 50 unfaithful responses produced by the \texttt{Llama-2-7B-Chat} model on the CLAPNQ dataset. We manually identified and categorized several patterns of unfaithfulness as follows.

\paragraph{Fabricated Hallucination.}
This error type refers to LLMs generating completely fabricated facts that are not derived from the provided context. Such unfaithfulness typically arises from retrieval failures where the retriever is unable to retrieve relevant information related to the question, or from the model relying solely on its parametric knowledge to generate an answer.
In our \benchmark, the documents within the context contain sufficient information to answer the questions. Therefore, this type of unfaithfulness typically stems from the latter scenario.
\paragraph{Incomplete Hallucination.}
This error type refers to the model generating content that is partially faithful, with the remainder of the content being fabricated. It generally occurs when the model captures only partial relevant information or when the context contains incomplete information. Consequently, the model may utilize its parametric knowledge to complete the response, leading to incomplete hallucinations.
\paragraph{Inconsistent Hallucination.}
This error type refers to the model generating content that contradicts the provided context. Typically, this stems from the model inaccurately synthesizing information from separate documents, resulting in content that is inconsistent with the provided context.

\section{Detailed Data Statistics of \ours}
\label{appendix:datasets}
We provide detailed dataset statistics of \benchmark in Table~\ref{tab:data_statistic}, covering the instance count, query sources, document sources, the average length of responses, as well as the average length of documents.
\begin{table*}[t]
\small
\centering
\resizebox{\linewidth}{!}{
\begin{tabular}{@{}lcccccc@{}}
\toprule
\textbf{Dataset} & \textbf{Number} & \textbf{Question Sources} & \textbf{Response Sources} & \textbf{Document Sources} & \textbf{Response Avg. Len.} & \textbf{Documents Avg. Len.} \\ 
\midrule
ExpertQA   & 527 & Domain Experts & Human \& Bing-Chat \& GPT-4 & Web Search & 134.6 & 617.0 \\
ELI5-WebGPT  & 271 & ELI5 & Human & Web Search & 106.0 & 268.7 \\
CLAPNQ & 300 & NQ & Human & Wikipedia & 51.7 & 632.2 \\
HAGRID & 496 & MIRACL & Human \& GPT-3.5 & Wikipedia & 41.15 & 302.2 \\
QuoteSum & 1,319 & NQ \& PAQ & Human & Wikipedia & 60.3 & 319.1 \\
\bottomrule
\end{tabular}
}
\caption{Statistics of datasets used in \benchmark.}
\vspace{-4mm}
\label{tab:data_statistic}
\end{table*}

\section{Details of Automatic Metrics}
\label{appendix:automatic_metrics}
We provide a detailed description of the evaluation metrics employed to assess the overall quality and faithfulness of the model-generated responses.
\paragraph{Overall Quality.}
To evaluate the overall quality of model-generated responses, we compare them to human-labeled answers. 
We utilize different automatic metrics tailored to the specific formats of the human-labeled responses across various datasets.
For ExpertQA, CLAPNQ, and HAGRID, we directly use the ROUGE-L score to measure the n-gram similarity between the model's responses and the human-labeled answers.
For ELI5-WebGPT, which typically features longer responses, we follow the methodology described in \citet{gao2023enabling} and use ChatGPT\footnote{We use gpt-3.5-turbo-1106 version.} to decompose the human answers into sub-claims. We then assess the overall quality by comparing the degree of entailment between the model's responses and these sub-claims. 
For QuoteSum, which is designed as a semi-extractive QA task, models are required to synthesize information from multiple sources while explicitly extracting factual spans. 
The human-written semi-extractive answers predominantly consist of quoted answers.
Following the evaluation protocol in \citet{schuster2024semqa}, answers are assessed for fluency and attribution preciseness.
Fluency is quantified using the ROUGE-L score. Preciseness is measured by calculating the normalized token-F1 score for each source separately and then averaging these scores across all sources, taking the maximum score per source across reference quoted answers. Finally, we compute the geometric mean of these scores, referred to as SEMQA, to reflect the overall answer quality.

\paragraph{Faithfulness.}
Faithfulness is the most critical evaluation dimension in \benchmark, assessing whether the model-generated response is fully supported by the provided documents. To evaluate faithfulness, we first segment the model-generated long-form response into statements by sentence boundaries\footnote{Specifically, we employ the NLTK library for this segmentation.}. We then assess the entailment of each sentence against the provided context using an NLI-based model. Specifically, we utilize \texttt{Bespoke-MiniCheck-7B}, the most advanced variant of MiniCheck \citep{tang2024minicheck} during the evaluation. The overall faithfulness score is calculated by the average score across all sentences in the response.

\section{Details of Human Evaluation}
\label{appendix:human_evaluation}
We recruited two annotators, each holding at least a bachelor's degree, to participate in the human evaluation.
The evaluation focuses on two key aspects: faithfulness and completeness of the responses.
For faithfulness, we selected the outputs from four systems' across the five datasets in \benchmark: GPT-4o, Llama-3.1-70B-Instruct, SFT-Llama-2-13B, and \ours-13B, resulting in a total of 200 generations.
Unlike the NLI-based model that outputs binary labels, we ask the annotators to perform a more fine-grained evaluation, determining whether each statement in the responses is fully supported, partially supported, or not supported by the provided context.
For completeness, annotators assessed whether the responses fully addressed the given questions and captured all relevant information in the provided context.
During the evaluation, both dimensions were rated using a 5-point Likert scale, capturing varying levels of faithfulness and completeness.

\section{Details of Implementation}
\label{appendix:implementation}

\subsection{Training Dataset}
As for the training dataset, we utilize the FRONT~\citep{huang2024learningfinegrainedgroundedcitations} dataset for training, selected for its diverse and high-quality retrieval documents refined through a rigorous data filtering process to ensure relevance to the questions. The dataset comprises 5,677 diverse user questions from the Natural Questions (NQ) dataset \citep{kwiatkowski2019natural}, designed for long-form responses, and 2,431 questions for short-form factoid answers. Specifically, we only use the long-form split for training the baselines and \ours.

\subsection{Prompts for Training and Evaluation}
\begin{figure*}
    \centering
    \footnotesize{
\begin{tikzpicture}

\begin{scope}[]

\node [anchor=south west] (n1) at (0, 0) {};

\node [anchor=north,rectangle,rounded corners=10pt,minimum height=1.7in,minimum width=6.0in, draw, line width=1.5pt, draw=black!60!white] (b1) at ([xshift=0em, yshift=-1.0em]n1.south) {};

\node [anchor=center,rectangle,rounded corners=5pt,minimum height=1em,minimum width=2em,fill=teal!30, draw=black!60!white, line width=1.3pt] (l1) at ([xshift=0em, yshift=0em]b1.north) {\small{(a) Prompt template used for training and evaluation}};

\node [anchor=north west, rectangle, rounded corners=2pt, minimum height=1em, minimum width=2em, text width=5.8in] (n21) at ([xshift=0.5em, yshift=-1em]b1.north west) {\texttt{\textbf{\#\#\# Instruction:} \\ Write an accurate, engaging, and concise answer to the given question using only the provided retrieval documents. Ensure the answer is well-grounded in the relevant information, disregarding irrelevant information in documents.}};
\node [anchor=north west,rectangle,rounded corners=2pt,minimum height=1em,minimum width=2em, text width=5.8in] (n22) at ([xshift=0em, yshift=-0.2em]n21.south west) {\texttt{\textbf{\#\#\# Input:} \\ }};
\node [anchor=north west,rectangle,rounded corners=2pt,minimum height=1em,minimum width=2em, text width=5.8in] (n23) at ([xshift=0em, yshift=-0.2em]n22.south west) {\texttt{\textbf{Question:} [Question]}};
\node [anchor=north west,rectangle,rounded corners=2pt,minimum height=1em,minimum width=2em, text width=5.8in] (n24) at ([xshift=0em, yshift=-0.2em]n23.south west) {\texttt{\textbf{Retrieved documents:} [Documents]}};
\node [anchor=north west,rectangle,rounded corners=2pt,minimum height=1em,minimum width=2em, text width=5.8in] (n25) at ([xshift=0em, yshift=-0.2em]n24.south west) {\texttt{\textbf{\#\#\# Response:}}};

\begin{pgfonlayer}{background}
\node [anchor=north,rectangle,rounded corners=3pt,minimum height=1.5in,minimum width=5.85in,fill=tiffanyblue!40] (bb) at ([xshift=0em, yshift=-0.2em]l1.south) {};

\end{pgfonlayer}

\end{scope}

\end{tikzpicture}
}
\caption{Illustration of the prompting used for training and evaluation.}
\label{fig:prompt_ablations}
\end{figure*}
The prompt template employed in both the training and evaluation phases is shown in Figure~\ref{fig:prompt_ablations}.
The prompt explicitly instructs the model to ensure the generation is well-grounded in the retrieved documents.
We use this template for training the SFT baseline and \ours, as well as for evaluation the SFT, prompting baseline, and \ours.

\subsection{Baselines}
\label{appendix:baseline_detail}
The detailed baseline implementation is as follows:

\paragraph{Prompting-based Method:}
For each question in \benchmark, the method involves directly prompting LLMs to generate responses based on the provided retrieved documents.
To comprehensively evaluate \benchmark, we employ both proprietary and open-source LLMs.
For proprietary LLMs, we use state-of-the-art models, including GPT-4o and its more efficient variant, GPT-4o-mini.
Regarding open-sourced LLMs, we cover a range of model sizes, from the most powerful, such as Llama-3.1-70B-Instruct\footnote{\url{https://huggingface.co/meta-llama/Meta-Llama-3.1-70B-Instruct}}, to smaller models like Mistral-NeMo-12B-Instruct\footnote{\url{https://huggingface.co/nvidia/Mistral-NeMo-12B-Instruct}} and Llama-3.1-8B-Instruct\footnote{\url{https://huggingface.co/meta-llama/Llama-3.1-8B-Instruct}}.
During the evaluation, we utilized the vLLM framework \citep{kwon2023efficient} for efficient inference and adopted a sampling decoding strategy for all models, setting the temperature to 1.0 and top-p to 0.95. The maximum generation length was set to 512 tokens.

\paragraph{Supervised Fine-Tuning:}
The method involves fine-tuning the model on retrieved documents to generate long-form responses to the given questions. In this way, models are trained to leverage contextual evidence to produce a well-grounded response.
Instead of fine-tuning using the existing responses from the FRONT dataset, which are generated by ChatGPT with specific instructions for attributed text generation~\citep{huang2024learningfinegrainedgroundedcitations} and may introduce additional bias, we employ the Llama-3.1-70B-Instruct model to generate faithful responses as the supervised outputs, chosen for its remarkable faithfulness in long-form QA.
During training, we use the Llama-2-7B and Llama-2-13B models. Specifically, we conduct full fine-tuning for three epochs using eight A100-80GB GPUs. The total batch size is set to 64, and the learning rate is maintained at 2e-5. The maximum input sequence length is set to 2,048 tokens. 
During decoding, we adopt the same decoding configuration as the prompting-based method.

\paragraph{RECOMP \citep{xu2023recomp}:}
The method involves training an abstractive summarization model to filter irrelevant information from the context.
Specifically, for each question and provided documents in FRONT, we first utilize GPT-3.5-turbo to summarize all the important information needed to answer the question. 
\begin{figure*}
    \centering
    \footnotesize{
\begin{tikzpicture}

\begin{scope}[]

\node [anchor=south west] (n1) at (0, 0) {};

\node [anchor=north,rectangle,rounded corners=10pt,minimum height=1.2in,minimum width=6.0in, draw, line width=1.5pt, draw=black!60!white] (b1) at ([xshift=0em, yshift=-1em]n1.south) {};

\node [anchor=center,rectangle,rounded corners=5pt,minimum height=1em,minimum width=2em,fill=teal!30, draw=black!60!white, line width=1.3pt] (l1) at ([xshift=0em, yshift=0em]b1.north) {\small{(a) Prompt template used for generating summarization}};

\node [anchor=north west, rectangle, rounded corners=2pt, minimum height=1em, minimum width=2em, text width=5.8in] (n21) at ([xshift=0.5em, yshift=-1em]b1.north west) {\texttt{\textbf{Instruction:} Summarize all the important information related to the question in the retrieved documents to answer the question:}};
\node [anchor=north west,rectangle,rounded corners=2pt,minimum height=1em,minimum width=2em, text width=5.8in] (n22) at ([xshift=0em, yshift=-0.2em]n21.south west) {\texttt{\textbf{Question:} [Question]}};
\node [anchor=north west,rectangle,rounded corners=2pt,minimum height=1em,minimum width=2em, text width=5.8in] (n23) at ([xshift=0em, yshift=-0.2em]n22.south west) {\texttt{\textbf{Retrieved documents:} [Documents]}};
\node [anchor=north west,rectangle,rounded corners=2pt,minimum height=1em,minimum width=2em, text width=5.8in] (n24) at ([xshift=0em, yshift=-0.2em]n23.south west) {\texttt{\textbf{Summary:}}};

\begin{pgfonlayer}{background}
\node [anchor=north,rectangle,rounded corners=3pt,minimum height=1.0in,minimum width=5.85in,fill=tiffanyblue!40] (bb) at ([xshift=0em, yshift=-0.2em]l1.south) {};

\end{pgfonlayer}

\end{scope}

\begin{scope}[]

\node [anchor=south west] (n1) at (0, -4) {};

\node [anchor=north,rectangle,rounded corners=10pt,minimum height=1.6in,minimum width=6.0in, draw, line width=1.5pt, draw=black!60!white] (b1) at ([xshift=0em, yshift=-1em]n1.south) {};

\node [anchor=center,rectangle,rounded corners=5pt,minimum height=1em,minimum width=2em,fill=teal!30, draw=black!60!white, line width=1.3pt] (l1) at ([xshift=0em, yshift=0em]b1.north) {\small{(b) Prompt template used for evaluation}};

\node [anchor=north west, rectangle, rounded corners=2pt, minimum height=1em, minimum width=2em, text width=5.8in] (n21) at ([xshift=0.5em, yshift=-1em]b1.north west) {\texttt{\textbf{\#\#\# Instruction:} \\ Write an accurate, engaging, and concise answer to the given question using only the provided documents. Ensure the answer is well-grounded in the provided information.}};
\node [anchor=north west,rectangle,rounded corners=2pt,minimum height=1em,minimum width=2em, text width=5.8in] (n22) at ([xshift=0em, yshift=-0.2em]n21.south west) {\texttt{\textbf{\#\#\# Input:} \\ }};
\node [anchor=north west,rectangle,rounded corners=2pt,minimum height=1em,minimum width=2em, text width=5.8in] (n23) at ([xshift=0em, yshift=-0.2em]n22.south west) {\texttt{\textbf{Question:} [Question]}};
\node [anchor=north west,rectangle,rounded corners=2pt,minimum height=1em,minimum width=2em, text width=5.8in] (n24) at ([xshift=0em, yshift=-0.2em]n23.south west) {\texttt{\textbf{Summarized documents:} [Documents]}};
\node [anchor=north west,rectangle,rounded corners=2pt,minimum height=1em,minimum width=2em, text width=5.8in] (n25) at ([xshift=0em, yshift=-0.2em]n24.south west) {\texttt{\textbf{\#\#\# Response:}}};

\begin{pgfonlayer}{background}
\node [anchor=north,rectangle,rounded corners=3pt,minimum height=1.4in,minimum width=5.85in,fill=tiffanyblue!40] (bb) at ([xshift=0em, yshift=-0.2em]l1.south) {};

\end{pgfonlayer}

\end{scope}

\end{tikzpicture}
}
\caption{Illustration of the prompting used for RECOMP.}
\label{fig:prompt_recomp}
\end{figure*}
Following the setting in \citet{xu2023recomp}, we train an abstractive summarization model, \eg, Flan-T5-Large \citep{chung2024scaling} using the distilled summarization dataset above. The summarization model is then employed to summarize all documents in \benchmark, ensuring that irrelevant information is filtered out to avoid distraction during faithful generation.
During the evaluation, we prepend the corresponding summarization to each question in \benchmark to guide the models in generating faithful long-form responses.
In particular, we use the chat version of Llama-2-7b and Llama-2-13b for evaluation.
The prompt templates used for both summarization and evaluation are presented in Figure~\ref{fig:prompt_recomp}.

\paragraph{Self-RAG \citep{asai2024selfrag}:}
The method trains an LLM that adaptively retrieves documents on-demand and generates and reflects on retrieved documents and its own generations using reflection tokens. These reflection tokens are categorized into retrieval and critique tokens to indicate the need for retrieval and the attributability of its generation, respectively.
Specifically, we directly utilized models provided by the authors, trained using Llama-2-7b\footnote{\url{https://huggingface.co/selfrag/selfrag_llama2_7b}} and Llama-2-13b\footnote{\url{https://huggingface.co/selfrag/selfrag_llama2_13b}}. During the evaluation, we employ the official prompt template designed for the ASQA task, using a maximum decoding length of 512 tokens.
We used the default configuration for long-form generation, with relevance, attribution, and overall completeness critique rewards each weighted at 1.0. During decoding, we adopt a beam width of 7 at each segment level.

\subsection{Implemantation for \ours}
\ours consists of three main components: unfaithful data augmentation, faithfulness-aware tuning, and self-induced decoding.

\paragraph{Unfaithful Data Augmentation:} 
In this stage, for each question and its corresponding documents in the FRONT dataset, we mask out the top 100 retrieval heads in both Llama-2-7B and 13B models. The responses generated from these models are categorized as unfaithful and are subsequently used for fine-tuning. This process yields 5,677 unfaithful responses for both the Llama-2-7B and Llama-2-13B models, respectively.

\paragraph{Faithfulness-Aware Tuning:} 
In this phase, we utilize the unfaithful responses generated during the data augmentation stage along with the faithful responses from the SFT baseline training data for fine-tuning.
Training is conducted on eight A100-80GB GPUs using Deepspeed~\citep{rasley2020deepspeed} stage 3 for efficient multi-GPU distribution, with training precision set to Bfloat16. We maintain a total batch size of 64, a learning rate of 2e-5, and a maximum input sequence length of 2,048 tokens. Both the 7B and 13B models are trained over 3 epochs.

\paragraph{Self-Induced Decoding:} We set $\alpha=0.2$ for both \ours-7B and \ours-13B models, with a decoding temperature of 1.0 and a top-p setting of 0.95. The same prompt template used for the SFT baseline, as depicted in Figure~\ref{fig:prompt_ablations}, is employed for both training and evaluation.

\section{Negative Sample Augmentation Strategies}
\label{appendix:negative_sample}
Due to our task scenario being retrieval-augmented LFQA, our goal is to generate unfaithful responses that are not faithful to the provided context. In LFQA, the main types of faithfulness hallucination include context inconsistency where generated responses directly contradict the context, mainly involving inconsistencies in entities and relationships \citep{mishra2024fine}, as well as fabric hallucinations that completely make things up. Therefore, we have designed three strategies for generating negative samples based on entity-replacement, relation-replacement, and prompting-based approaches.

\paragraph{Entity and Relation Replacement.}
This strategy involves replacing key entities and relations in the provided faithful response with other plausible but contextually irrelevant alternatives. We adopt the methodologies outlined in \citet{mishra2024fine}, instructing powerful LLMs, \eg, Llama-3.1-70B-Instruct, to execute entity or relation replacements. This process is designed to generate a diverse array of unfaithfulness error types, resulting in synthetic responses that are naturally contradictory to the contextual information.

\paragraph{Prompting-based.}

We employ the method in \citet{li2023halueval} for generating unfaithful samples.
Specifically, we directly feed the complete instruction shown in Figure~\ref{fig:prompt_negative} into the corresponding model (Llama-2-7B-Chat and Llama-2-13B-Chat) to generate a response inconsistent with the context.
In addition to instructing the model to produce a response inconsistent with the context, we also provide a faithful answer as a reference. This allows the model to generate diverse and multi-faceted unfaithful responses for each question.
Upon manual review, it was found that the primary error type in unfaithful responses generated by this method focuses on fabric hallucination, which involves the model generating responses based on parameterized knowledge that is unrelated to the given context.
\begin{figure*}
    \centering
    \footnotesize{
\begin{tikzpicture}

\begin{scope}[]

\node [anchor=south west] (n1) at (0, 0) {};
\node [anchor=north,rectangle,rounded corners=10pt,minimum height=1.9in,minimum width=6.0in, draw, line width=1.5pt, draw=black!60!white] (b1) at ([xshift=0em, yshift=-1.0em]n1.south) {};

\node [anchor=center,rectangle,rounded corners=5pt,minimum height=1em,minimum width=2em,fill=teal!30, draw=black!60!white, line width=1.3pt] (l1) at ([xshift=0em, yshift=0em]b1.north) {\small{(a) Prompt template used for training and evaluation}};

\node [anchor=north west, rectangle, rounded corners=2pt, minimum height=1em, minimum width=2em, text width=5.8in] (n21) at ([xshift=0.5em, yshift=-1em]b1.north west) {\texttt{\textbf{\#\#\# Instruction:} \\ I want you act as a hallucination answer generator. Given a question, provided documents, and a relatively faithful answer, write a unfaithful answer that is not grounded in the provided documents.}};
\node [anchor=north west,rectangle,rounded corners=2pt,minimum height=1em,minimum width=2em, text width=5.8in] (n22) at ([xshift=0em, yshift=-0.2em]n21.south west) {\texttt{\textbf{\#\#\# Input:} \\ }};
\node [anchor=north west,rectangle,rounded corners=2pt,minimum height=1em,minimum width=2em, text width=5.8in] (n23) at ([xshift=0em, yshift=-0.2em]n22.south west) {\texttt{\textbf{Question:} [Question]}};
\node [anchor=north west,rectangle,rounded corners=2pt,minimum height=1em,minimum width=2em, text width=5.8in] (n24) at ([xshift=0em, yshift=-0.2em]n23.south west) {\texttt{\textbf{Retrieved documents:} [Documents]}};
\node [anchor=north west,rectangle,rounded corners=2pt,minimum height=1em,minimum width=2em, text width=5.8in] (n25) at ([xshift=0em, yshift=-0.2em]n24.south west) {\texttt{\textbf{Faithful answer:} [Faithful Answer]}};
\node [anchor=north west,rectangle,rounded corners=2pt,minimum height=1em,minimum width=2em, text width=5.8in] (n26) at ([xshift=0em, yshift=-0.2em]n25.south west) {\texttt{\textbf{\#\#\# Response:}}};

\begin{pgfonlayer}{background}
\node [anchor=north,rectangle,rounded corners=3pt,minimum height=1.7in,minimum width=5.85in,fill=tiffanyblue!40] (bb) at ([xshift=0em, yshift=-0.2em]l1.south) {};

\end{pgfonlayer}

\end{scope}

\end{tikzpicture}
}
\caption{Illustration of the prompting used for generating unfaithful responses.}
\label{fig:prompt_negative}
\end{figure*}

\section{Additional Analysis Experiments}
\label{appendix:analysis}
\paragraph{Negative Samples from Different Models.}
To explore the impact of negative samples induced from different models on faithfulness, we utilized the Llama-2-7B and Llama-2-13B models to generate negative samples by masking out their own top 100 retrieval heads. These self-induced negative samples were then employed to train each model separately. As shown in Figure~\ref{fig:ablation_self_induced} (a), models trained with their own negative samples, e.g., Llama-2-7B with samples induced by itself (7B-Induced) and Llama-2-13B with samples induced by itself (13B-Induced), demonstrated superior performance compared to models trained with negative samples induced by the other model. This indicates that self-induced negative samples may be more closely aligned with each model's specific error tendencies, thereby providing more effective training for improving faithfulness.

\paragraph{Impact of the Number of Masked Retrieval heads.}
We investigate the impact of the number of masked retrieval heads on model faithfulness. We conduct an ablation study to mask the top 50, 100, and 150 retrieval heads of Llama-2-7B to generate respective sets of unfaithful samples. As shown in Figure~\ref{fig:ablation_self_induced} (b), increasing the number of masked retrieval heads from 50 to 100 led to an improvement in the model's faithfulness, suggesting that masking more retrieval heads can enhance faithfulness by generating more challenging negative samples. However, further increasing the number of masked heads to 150 resulted in a decline in faithfulness. We hypothesize that excessively masking retrieval heads degrades the quality of the generated unfaithful samples, introducing issues such as reduced coherence. This degradation could hinder the model's ability to effectively learn from these unfaithful samples, ultimately impacting the training process negatively.

\definecolor{red}{RGB}{172,21,28}
\definecolor{blue}{RGB}{39,89,167}
\definecolor{red1}{RGB}{203,104,104}
\definecolor{blue1}{RGB}{104,155,203}
\definecolor{color1}{RGB}{48,75,116}
\definecolor{color2}{RGB}{108,149,183}

\begin{figure}[t]
    \centering
    \vspace{1.0mm}
\begin{tikzpicture}
    \scriptsize{
      \begin{axis}[
    at={(-4.5em,-5em)},
    anchor=south west,
    ymajorgrids,
    grid style=dashed,
    legend style={at={(-0.1,0.65)}, anchor=south west},
    legend cell align={left},
    ybar,
    enlarge x limits=0.5,
    xtick align=inside,
    height=.26\textwidth,
    width=.27\textwidth,
    bar width=1.3em,
    xlabel={\scriptsize{(a) Model Sizes}},
    xlabel style={scale=1.2, yshift=0.8em, xshift=0.1em},
    ylabel=\footnotesize{\scriptsize Avg. FaithScore},
    ylabel style={scale=1.2, yshift=0.5em},
    symbolic x coords={{1}, {2}},
    xtick=data,
    ymin=0.795,
    ymax=0.865,
    ytick={0.80, 0.81, 0.82, 0.83,0.84, 0.85, 0.86},
    nodes near coords align={vertical},
    xticklabels={Llama-2-7B, Llama-2-13B},
    ylabel style={yshift=-2em},
    yticklabel style={/pgf/number format/fixed,/pgf/number format/fixed zerofill,/pgf/number format/precision=2,rotate=0,scale=1.0},
    legend style={yshift=0.2em,xshift=4.2em,font={\tiny},cells={anchor=west},fill opacity=0.8, scale=1.0, legend columns=1}
    ]
    \addplot[fill=color1, draw=color1, area legend] coordinates {({1},0.8235) ({2},0.8209)};
    \addlegendentry{\scalebox{0.8}{{7B-Induced}}}
    \addplot[fill=color2, draw=color2, area legend] coordinates {({1},0.8099) ({2},0.8377)};
    \addlegendentry{\scalebox{0.8}{{13B-Induced}}}
  \end{axis}
}  
    \begin{axis}
    [
        anchor=north west,
        at={(11.5em,5.5em)},
        ymajorgrids,
        xmajorgrids,
        grid style=dashed,
        width=.27\textwidth,
        height=.26\textwidth,
        yticklabel style={/pgf/number format/precision=2,/pgf/number format/fixed zerofill,scale=1.0},
        xmax=1300,
        xmin=300,
        ymin=0.785,
        ymax=0.835,
        xtick={400,800,1200},
        xticklabels={50,100,150},
        ytick={0.78,0.79,0.80,0.81,0.82,0.83},
        xlabel={\scriptsize{(b) Number of masked heads}},
        xlabel style={scale=1.2, yshift=0.8em, xshift=0.1em},
        ylabel=\footnotesize{\scriptsize Avg. FaithScore},
        ylabel style={yshift=-1.9em, scale=1.2},
legend style={at={(0.88,0.87)}, anchor=north east, font={\tiny}, cells={anchor=west}, fill opacity=0.8, scale=1.0}
        ]

\addplot[red,mark=pentagon*,,mark size=1.8pt,thick,mark options={fill=white,draw=red,line width=1pt}] coordinates {(400,0.8115) (800,0.8235) (1200,0.7986)};
    \end{axis} 
\end{tikzpicture}
    \caption{
    Analysis of the impact of self-induced negative samples and the number of masked retrieval heads.
    }
    \label{fig:ablation_self_induced}
\end{figure}
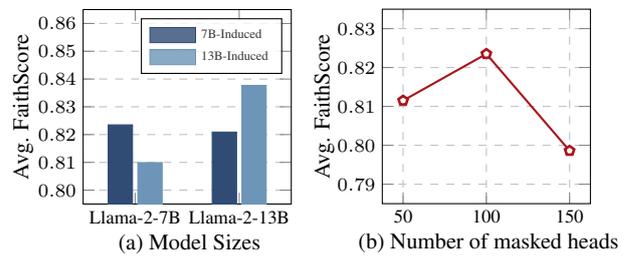

\section{Case Study}
\label{appendix:case_study}

\begin{table*}[h]
    \centering
    \small
    \begin{tabular}{>{\raggedright\arraybackslash\tt}p{0.98\textwidth}<{}}
        \toprule
            \#\#\# \textbf{Instruction:} \\ 
            Write an accurate, engaging, and concise answer to the given question using only the provided retrieval documents. Ensure the answer is well-grounded in the relevant information, disregarding irrelevant information in documents. \\
            \#\#\# \textbf{Input:} \\ 
            \textbf{Question:} Difference between Russian Blue and British Blue cat \\
            \\
            \vspace{-1em}
            \textbf{Retrieved documents:} \\
            Document [1]: The Russian Blue is a cat breed that comes in colors varying from a light shimmering silver to a darker, slate grey. They develop close bonds with their owners and are sought out as pets due to their personalities, beauty and coat. It is their short, dense coat which has been the hallmark of the Russian breed for more than a century. The dense coat stands out from the body and one can draw patterns in the coat that will stay until one smoothes them out again. They are also considered to be hypoallergenic.\\
            Document [2]: The Russian Blue has bright green eyes, pinkish lavender or mauve paws, two layers of short thick fur, and a blue - grey coat. The color is a bluish - gray that is the dilute expression of the black gene. However, as dilute genes are recessive (``d'') and each parent will have a set of two recessive genes (``dd'') two non-CPC Russian Blues will always produce a blue cat. Due to the breeding with Siamese after World War II, there are colorpoint genes floating around. If two carriers are bred together, then they will produce a litter of mixed colors -- solid blue or white with blue like a Siamese. People call these CPC cats ``colorpoint'' , ``whites'' or ``pointed'' Russians. In most registries, one can not register, breed or show a colorpoint Russian. \\
            Document [3]: The coat is known as a ``double coat'', with the undercoat being soft, downy and equal in length to the guard hairs, which are an even blue with silver tips. However, the tail may have a few very dull, almost unnoticeable stripes. The coat is described as thick, plush and soft to the touch. The feeling is softer than the softest silk. The silver tips give the coat a shimmering appearance. Its eyes are almost always a dark and vivid green. Any white patches of fur or yellow eyes in adulthood are seen as flaws in show cats. Russian Blues should not be confused with British Blues (which are not a distinct breed, but rather a British Shorthair with a blue coat as the British Shorthair breed itself comes in a wide variety of colors and patterns), nor the Chartreux or Korat which are two other naturally occurring breeds of blue cats, although they have similar traits. \\
            Document [4]: Russian Blue 1 year old Russian Blue male (American style) Other names Archangel Blue, Archangel Cat Origin Russia Breed standards CFA standard FIFe standard TICA standard AACE standard ACF standard ACFA / CAA standard CCA - AFC standard GCCF standard Notes The ACF and GCCF also recognize Russian Blues in white and black, however the CFA does not. In addition, ACFA recognizes Russian Shorthairs in white, black, and blue. Domestic cat (Felis catus) \\
            Document [5]: The Russian Blue is a naturally occurring breed that may have originated in the port of Arkhangel'sk in Russia. They are also sometimes called Archangel Blues. It is believed that sailors took Russian Blues from the Archangel Isles to Great Britain and Northern Europe in the 1860s. The first recorded appearance outside of Russia was in 1875 at The Crystal Palace in England as the Archangel Cat. The Russian Blue competed in a class including all other blue cats until 1912, when it was given its own class. The breed was developed mainly in England and Scandinavia until after World War II. \\
            \\
            \vspace{-1em}
            \#\#\# \textbf{Response:} \\
            \vspace{-1em}
            {\color{blue} The Russian Blue cat is known for its bright green eyes, pinkish lavender or mauve paws, two layers of short thick fur, and a blue-grey coat. The color of the Russian Blue is a bluish-gray, which is the dilute expression of the black gene. On the other hand, British Blues are not a distinct breed but rather a British Shorthair with a blue coat. The British Shorthair breed itself comes in a wide variety of colors and patterns, unlike the Russian Blue, which is a specific breed known for its unique coat and characteristics.}\\
        
        \bottomrule
    \end{tabular}
    \caption{Case study of how \ours helps the model to generate the faithful response from CLAPNQ dataset.}
    \label{tab:example_clapnq}
\end{table*}
We provide a case study in Table~\ref{tab:example_clapnq}. It can be seen that compared to the response generated by GPT-4o-mini in Table~\ref{tab:example_lfqa}, which contains hallucinated facts and inaccurate information synthesizing, our model demonstrates higher faithfulness and well-grounded in the provided documents.

\end{document}